\documentclass[11pt]{article}

\usepackage[margin=1in]{geometry}
\usepackage[utf8]{inputenc}
\usepackage[T1]{fontenc}
\usepackage{lmodern}
\usepackage{amsmath,amssymb,amsfonts}
\usepackage{graphicx}
\usepackage{booktabs}
\usepackage{xcolor}
\usepackage{algorithm}
\usepackage{algorithmic}
\usepackage{caption}
\usepackage{subcaption}
\usepackage{multirow}
\usepackage{microtype}
\usepackage{natbib}
\usepackage{hyperref}
\usepackage{url}

\hypersetup{
    colorlinks=true,
    linkcolor=blue,
    citecolor=blue,
    urlcolor=blue
}

\newcommand{\dmodel}{d_{\text{model}}}
\newcommand{\dselect}{d_{\text{select}}}

\newcommand{\dhead}{d_{\text{head}}}
\newcommand{\softmax}{\text{softmax}}
\newcommand{\BigO}{\mathcal{O}}

\title{Thin Keys, Full Values: \\
Reducing KV Cache via Low-Dimensional Attention Selection}

\author{%
  Hengshuai Yao\textsuperscript{1,2} \quad Xing Chen\textsuperscript{1} \quad Ahmed Murtadha\textsuperscript{1} \quad Guan Wang\textsuperscript{1} \\
  \textsuperscript{1}Sapient Intelligence \\
  \textsuperscript{2}Department of Computing Science, University of Alberta \\
  \texttt{hengshu1@ualberta.ca, raincchio@gmail.com,} \\
  \texttt{murtadha20@gmail.com, imonenext@gmail.com}%
}

\begin{document}

\maketitle

\begin{abstract}
Standard transformer attention uses identical dimensionality for queries, keys, and values, yet these components serve different roles: queries and keys produce scalar attention weights (\emph{selection}), while values carry rich representations (\emph{value transfer}). We show that selection requires only $\BigO(\log N)$ dimensions to distinguish among $N$ relevant token categories (e.g., syntactic roles, semantic clusters, positional patterns)---far fewer than value transfer needs.

We introduce \textbf{factored keys}, which exploit this asymmetry to physically shrink the KV cache of \emph{any pretrained model without retraining from scratch}---unlike GQA and MLA, which must be designed into the architecture before pretraining. We factorize each key projection $W_K \approx A_{d \times r} B_{r \times d}$ via truncated SVD (where $r = \dselect$), set $W_K' = A$ as the new key projection producing compact $r$-dimensional keys for the cache, and absorb $B^\top$ into the query projection ($W_Q' = W_Q B^\top$) at zero cost---since queries are never cached. At 7B scale, training from scratch with $\dselect = \dmodel/4$ matches full-attention perplexity ($9.24$ vs $9.25$ PPL after 20B tokens, mean over 2 seeds) while using 12\% fewer parameters and training 8\% faster. For existing models, SVD + QK fine-tuning (3 epochs, $<$1\% of pretraining data) achieves 75\% key cache savings at $\sim$2\% quality cost on both GPT-2 and Mistral-7B. The approach composes with GQA and quantization for up to 16$\times$ combined key cache compression. For a 7B model serving 128K context, factored keys save 25\,GB of KV cache per user, enabling $\sim$60\% more concurrent users on identical hardware.
\end{abstract}

% ============================================================
% 1. INTRODUCTION
% ============================================================
\section{Introduction}

The self-attention mechanism \citep{vaswani2017attention} projects inputs into queries, keys, and values ($Q$, $K$, $V$) and computes $\softmax(QK^\top\!/\!\sqrt{d_k})\,V$. In virtually all modern transformers---GPT \citep{radford2018improving}, LLaMA \citep{touvron2023llama}, Mistral \citep{jiang2023mistral}---the projection dimensionality is identical: $d_q = d_k = d_v = \dmodel$. This symmetry is a design convention, not a necessity.

As models scale to longer contexts, the KV cache becomes the dominant memory bottleneck during autoregressive inference: a 7B model serving 128K context requires over 60\,GB of KV cache per user. Existing approaches to this bottleneck reduce the \emph{number} of KV heads---Multi-Query Attention \citep{shazeer2019fast} and Grouped-Query Attention \citep{ainslie2023gqa}---or compress the \emph{joint} KV state into a low-rank latent. These methods have been widely adopted (GQA in LLaMA-2 \citep{touvron2023llama} onward; MLA in DeepSeek-V2 \citep{liu2024deepseek}), but they all maintain $d_k = \dhead$ within each head's attention computation.

We take a complementary approach, motivated by a structural observation: attention performs two functionally distinct operations with fundamentally different complexity requirements. (1)~\textbf{Selection} ($QK^\top$): determining \emph{which} tokens are relevant via scalar attention weights. (2)~\textbf{Value transfer} ($\text{attn} \cdot V$): aggregating information from selected tokens. Our key insight is that selection is a \emph{ranking} problem, not a representation problem: by the Johnson--Lindenstrauss lemma \citep{johnson1984extensions}, distinguishing among $N$ items in a dot-product space requires only $\BigO(\log N)$ dimensions. Value transfer, by contrast, must preserve the full representational capacity of the model. Recent theoretical work on KV cache compression lower bounds provides independent support: \citet{compressionbarriers2025} show that any attention-based token generation algorithm requires $\Theta(nd)$ space with $d = \Omega(\log n)$, establishing that the KV cache cannot be compressed below $\log n$ dimensions per token without loss. This asymmetry implies that queries and keys are drastically overparameterized at $d_k = \dmodel$.

We propose \textbf{asymmetric attention}, where queries and keys project to $\dselect \ll \dmodel$ while values retain full dimensionality. The attention computation is unchanged---$QK^\top / \sqrt{d_{\text{head}}^{QK}}$ produces scalar weights applied to $V$ of any dimensionality. This yields three benefits: (i)~parameter reduction (up to 75\% for QK when $\dselect = \dmodel/4$), (ii)~KV cache reduction (the key cache stores $\dselect$-dimensional vectors, saving 25\,GB per user at 128K context for 7B models), and (iii)~attention compute reduction ($\BigO(n^2 \cdot \dselect)$ instead of $\BigO(n^2 \cdot \dmodel)$). Unlike GQA or MLA, this approach reduces the \emph{per-head dimensionality} of keys while keeping the number of heads and value dimensionality intact---and composes with both for further savings.

A central contribution of this work is that asymmetric attention can be applied to \emph{any existing pretrained model} without retraining from scratch---in contrast to GQA and MLA, which must be built into the architecture before pretraining or require expensive conversion \citep{mha2mla2025}. We introduce \textbf{factored keys}, an inference primitive that physically shrinks the key cache of already-deployed models: factorize each layer's $W_K \approx A_{d \times r} B_{r \times d}$ via truncated SVD, use $A$ as the new key projection producing compact $r$-dimensional cached keys, and absorb $B^\top$ into the query projection at zero memory cost---exploiting the fundamental asymmetry that keys accumulate in memory while queries are ephemeral. With optional lightweight QK fine-tuning (3 epochs on $<$1\% of pretraining data), this recovers nearly all quality. This is related to recent work on low-rank KV cache compression \citep{lrqk2025, svdllm2025, loki2024}, but differs in a key respect: we compress \emph{only} keys and absorb the cost into queries, which are never cached. A striking finding is that keys are far more compressible than queries: at rank 192 on GPT-2, $K$-only SVD degrades by 26\% while $Q$-only degrades by 181\%---a 7$\times$ asymmetry consistent with recent observations that key vectors lie in a significantly lower-dimensional space \citep{loki2024}.

We validate across nine experiments from algorithmic tasks to 7B-scale training (Experiments~1--4 and 6 appear in Sections~\ref{app:small_scale}--\ref{app:arch_125m}; the four largest-scale experiments are presented first). Controlled experiments confirm that positional selection requires only 1 dimension per head, and content-based selection requires $\log_2 N$ dimensions (Section~\ref{app:small_scale}). At 7B scale, training from scratch with $\dselect = \dmodel/4$ matches full-attention perplexity after 20B tokens while using 12\% fewer parameters and training 8\% faster. For existing models, SVD + QK fine-tuning on Mistral-7B achieves 75\% key cache savings at $\sim$2\% quality cost.

% ============================================================
% 2. METHOD
% ============================================================
\section{Method}

\subsection{Asymmetric Attention}

In multi-head attention with $h$ heads, standard transformers set $\dhead = \dmodel / h$ for all of $Q$, $K$, and $V$. We decouple this by introducing $\dselect$: the per-head QK dimension becomes $d_{\text{head}}^{QK} = \dselect / h$ while the value dimension stays $d_{\text{head}}^{V} = \dmodel / h$. For each head $i$:
\begin{align}
Q_i &= XW_Q^{(i)}, \quad W_Q^{(i)} \in \mathbb{R}^{\dmodel \times d_{\text{head}}^{QK}} \\
K_i &= XW_K^{(i)}, \quad W_K^{(i)} \in \mathbb{R}^{\dmodel \times d_{\text{head}}^{QK}} \\
V_i &= XW_V^{(i)}, \quad W_V^{(i)} \in \mathbb{R}^{\dmodel \times d_{\text{head}}^{V}} \\
\text{head}_i &= \softmax\!\left(\frac{Q_i K_i^\top}{\sqrt{d_{\text{head}}^{QK}}}\right) V_i
\end{align}

No projection is needed between attention weights and value aggregation: the weights $\in \mathbb{R}^{n \times n}$ are scalars regardless of $d_{\text{head}}^{QK}$. When $\dselect = \dmodel$, this reduces to standard multi-head attention.

\subsection{Theoretical Motivation}\label{sec:theory}

We argue that selection requires $\BigO(\log N)$ dimensions where $N$ is the number of distinct patterns the attention mechanism must distinguish. The attention weight $\alpha_{ij} \propto \exp(q_i^\top k_j / \sqrt{d_{\text{head}}^{QK}})$ assigns scalar relevance scores---a ranking problem. By the Johnson--Lindenstrauss lemma, $N$ points can be embedded into $\BigO(\log N)$ dimensions while preserving pairwise distances, so $\BigO(\log N)$ dimensions suffice to maintain the relative ordering of dot-product similarities. In language modeling, the effective $N$ is the number of distinct selection patterns (syntactic roles, semantic clusters, recency patterns)---empirically in the hundreds, much smaller than vocabulary size, suggesting $d_{\text{head}}^{QK} \approx 10$--$20$ dimensions per head. Value transfer, by contrast, must preserve the complete representational content across all $\dmodel$ dimensions.

\subsection{Factored Keys: An Inference Primitive for Pretrained Models}

Given a pretrained $W_K \in \mathbb{R}^{\dmodel \times \dmodel}$, we seek a thin-key factorization of the form:
\begin{equation}
\underbrace{
\begin{bmatrix} & & & \\ \\ \quad & W_K & \quad \\ \\ & & & \phantom{x} \end{bmatrix}
}_{d \times d}
\;\approx\;
\underbrace{
\begin{bmatrix} \\ \\ W_K' \\ \\ \phantom{x} \end{bmatrix}
}_{\text{\textcolor{red}{\textbf{thin keys}}}\; (d \times r)}
\;
\underbrace{
\begin{bmatrix} & B & \end{bmatrix}
}_{r \times d \;\text{(absorbed into } W_Q\text{)}}
\end{equation}
Conveniently, truncated SVD provides exactly this factorization, with $r = \dselect \ll \dmodel$. We then replace the original projections:
\begin{align}
W_K' &= A = U_r \Sigma_r \in \mathbb{R}^{\dmodel \times r} & &\text{(\textbf{thin key} projection --- cached)} \label{eq:thin_key} \\
W_Q' &= W_Q B^\top = W_Q V_r \in \mathbb{R}^{\dmodel \times r} & &\text{(absorbed query projection --- ephemeral)} \label{eq:absorbed_query}
\end{align}
Keys become $k_j' = x_j W_K' \in \mathbb{R}^r$---the ``thin keys'' stored in the cache at a fraction of the original size. Queries become $q_i' = x_i W_Q' \in \mathbb{R}^r$, preserving attention scores exactly: $q_i' k_j'^\top = x_i W_Q B^\top A^\top x_j^\top = x_i W_Q W_K^\top x_j^\top = q_i k_j^\top$. Computing $W_Q'$ is a one-time matrix multiplication---no training or fine-tuning is required. This is the beauty of SVD: we obtain thin keys from any pretrained full-key model \emph{for free}, simply by repartitioning the existing weights. The result: key cache entries shrink from $\dmodel$ to $r$ dimensions while all compression cost is absorbed by the ephemeral query side (queries are computed fresh each step and never cached).

Crucially, \emph{nothing else in the network changes}---$W_V$, $W_O$, the FFN, and all normalization layers remain untouched. The only modification is replacing $W_K$ and $W_Q$ with their thin counterparts, computed from a single SVD. Despite this simplicity, training from scratch with thin keys at 7B scale achieves 9.2 vs 9.3 validation PPL after 20B tokens---matching full attention---while using 12\% fewer parameters and training 8\% faster (Experiments~7--7b, Tables~\ref{tab:7b_scratch} and~\ref{tab:7b_extended}).

This zero-cost property distinguishes factored keys from all existing KV compression methods. GQA \citep{ainslie2023gqa} and MLA \citep{liu2024deepseek} must be designed into the architecture before pretraining---they cannot be applied to already-deployed models. LRQK \citep{lrqk2025} performs QK decomposition at prefill time, adding latency to every inference call. SVD-LLM \citep{svdllm2025} requires calibration data and a whitening step to guide truncation. By contrast, factored keys require only a one-time offline SVD per layer---no calibration data, no prefill overhead, no retraining---and the resulting model runs with standard attention kernels. We validate this in Experiments~5 and~8: zero-cost SVD at rank $\dmodel/2$ immediately achieves 50\% key cache savings at +2\% PPL with no fine-tuning at all. To compress more aggressively, optional lightweight QK fine-tuning (3 epochs) recovers quality at lower ranks, enabling 75\% savings at rank $\dmodel/4$ with only +2\% PPL (Tables~\ref{tab:svd_finetune} and~\ref{tab:mistral7b}), with decode throughput gains increasing with batch size as the KV cache fraction of memory bandwidth grows (Table~\ref{tab:decode_throughput}).

\subsection{KV Cache Implications}

During autoregressive generation, the KV cache for $L$ layers at context length $n$ is:
\begin{align}
\text{Standard KV cache} &= 2 \cdot n \cdot \dmodel \cdot L \cdot b \text{ bytes} \\
\text{Asymmetric KV cache} &= n \cdot (\dselect + \dmodel) \cdot L \cdot b \text{ bytes}
\end{align}
where $b$ is bytes per parameter. With $\dselect = \dmodel / 4$, the K cache shrinks by 75\%, yielding 37.5\% total KV cache reduction.

% ============================================================
% 3. EXPERIMENTS
% ============================================================
\section{Experiments}

We validate asymmetric attention through nine experiments of increasing complexity. Controlled experiments on algorithmic tasks and small-scale language modeling (Experiments~1--4) are in Section~\ref{app:small_scale}; architecture generalization at 125M scale (Experiment~6) is in Section~\ref{app:arch_125m}. We first present the four largest-scale experiments: post-training SVD compression of GPT-2 (Experiment~5), from-scratch training at 7B scale (Experiments~7 and~7b), and SVD + fine-tuning of Mistral-7B (Experiment~8).

\subsection{Experiment 5: Post-Training Compression of GPT-2}

\paragraph{Setup.}
We apply SVD compression to pretrained GPT-2 (124M parameters, $\dmodel = 768$, 12 heads, 12 layers) without retraining, testing three modes: both $W_Q$ and $W_K$, $K$-only, and $Q$-only.

\paragraph{Results.}
Table~\ref{tab:svd} reveals a striking asymmetry. Compressing both $W_Q$ and $W_K$ is catastrophic---errors compound through the softmax. However, $K$-only compression is far more forgiving: rank 384 ($\dmodel/2$) incurs only 2.0\% degradation. $K$ is substantially more compressible than $Q$ at every rank: at rank 192, $K$-only degrades by 26\% while $Q$-only degrades by 181\%---a 7$\times$ asymmetry.

\begin{table}[h]
\centering
\caption{SVD compression of GPT-2 projections. Baseline PPL: 24.91. $\Delta$ is relative PPL increase.}
\label{tab:svd}
\begin{tabular}{rrrrr}
\toprule
Rank $r$ & $r$/head & Both $Q{+}K$ & $K$-only & $Q$-only \\
\midrule
128 & 10 & 67,132 & 57.37 (+130\%) & 149.04 (+498\%) \\
192 & 16 & 90.95 (+265\%) & 31.32 (+26\%) & 70.05 (+181\%) \\
256 & 21 & 46.15 (+85\%) & 27.49 (+10\%) & 40.42 (+62\%) \\
384 & 32 & --- & 25.40 (+2.0\%) & 27.58 (+11\%) \\
512 & 42 & 25.69 (+3.1\%) & 25.23 (+1.3\%) & 25.47 (+2.2\%) \\
\bottomrule
\end{tabular}
\end{table}

\paragraph{Deployment via factored keys.}
The SVD factorization provides a direct path to KV cache reduction: the key cache stores $r$-dimensional vectors while $B^\top$ is absorbed into the query projection. Our $K$-only PPL measurements directly reflect deployment performance: rank 384 ($\dmodel/2$) saves 50\% of the key cache (25\% total KV) at +2.0\% PPL; rank 512 saves 33\% of keys at +1.3\%.

\paragraph{Recovery via QK fine-tuning.}
We compress $W_K$ via SVD, then fine-tune only QK projections ($\sim$21M of 124M parameters) on WikiText-103 for 3 epochs over 10M tokens. Table~\ref{tab:svd_finetune} shows that fine-tuning nearly eliminates the SVD quality loss: at rank 192 ($\dmodel/4$), the gap shrinks from +27.6\% to just +1.8\% relative to an identically fine-tuned control.

\begin{table}[h]
\centering
\caption{SVD compression + QK fine-tuning on WikiText-103. The ``vs control'' column measures the residual gap relative to the uncompressed model after both receive identical fine-tuning.}
\label{tab:svd_finetune}
\begin{tabular}{rrrrrr}
\toprule
Rank $r$ & Before FT & After FT & Control & vs Control & K cache saved \\
\midrule
768 (none) & 29.51 & 19.07 & --- & baseline & 0\% \\
384 ($\dmodel/2$) & 30.15 (+2.2\%) & 19.14 & 19.07 & +0.4\% & 50\% \\
256 ($\dmodel/3$) & 32.61 (+10.5\%) & 19.28 & 19.07 & +1.1\% & 67\% \\
192 ($\dmodel/4$) & 37.64 (+27.6\%) & 19.42 & 19.07 & +1.8\% & 75\% \\
128 ($\dmodel/6$) & 67.26 (+128\%) & 19.77 & 19.07 & +3.7\% & 83\% \\
\bottomrule
\end{tabular}
\end{table}

\subsection{Experiment 7: From-Scratch Training at 7B Scale}

\paragraph{Setup.}
We train two LLaMA-7B models from random initialization on OpenWebText (2B tokens): a full-attention baseline and a thin-keys variant with $\dselect = 1024$ ($\dmodel/4$). Both use $\dmodel = 4096$, 32 heads, 32 layers, $d_{\text{ff}} = 11008$, with identical hyperparameters. We run each configuration with two random seeds and report mean $\pm$ std.

\begin{table}[h]
\centering
\caption{7B LLaMA trained from scratch on OpenWebText (2B tokens), mean $\pm$ std over 2 seeds.}
\label{tab:7b_scratch}
\begin{tabular}{lrr}
\toprule
 & Full Attention & Thin Keys ($\dselect = \dmodel/4$) \\
\midrule
Parameters & 6.74B & 5.93B ($-$12\%) \\
OWT val PPL & $13.82 \pm 0.09$ & $\mathbf{13.12 \pm 0.04}$ ($-$5.1\%) \\
WT103 val PPL & $20.79 \pm 0.35$ & $\mathbf{19.60 \pm 0.46}$ ($-$5.7\%) \\
Wall-clock time & 25.9h & 23.4h ($-$9.4\%) \\
\bottomrule
\end{tabular}
\end{table}

\paragraph{Results.}
Thin keys \emph{outperform} full attention at 7B scale with 2B tokens (Table~\ref{tab:7b_scratch}): 5.1\% better OWT perplexity while using 12\% fewer parameters and training 9.4\% faster. This inverts the $\sim$4\% cost observed at smaller scales (Sections~\ref{app:small_scale},~\ref{app:arch_125m}), consistent with a regularization effect in the overparameterized regime (tokens-to-parameters ratio $\sim$0.3). The training trajectory (Figure~\ref{fig:7b_trajectory}) shows thin keys leading at every checkpoint.

\begin{figure}[h]
\centering
\begin{tabular}{cc}
\small
\begin{tabular}{rcc}
\toprule
Step & Full & Thin \\
\midrule
5K  & 26.62 & 25.05 \\
10K & 20.81 & 19.21 \\
15K & 17.68 & 16.51 \\
20K & 15.66 & 14.74 \\
25K & 14.37 & 13.59 \\
30K & 13.79 & 13.12 \\
\bottomrule
\end{tabular}
&
\small
\begin{tabular}{rcc}
\toprule
Step & Full & Thin \\
\midrule
5K  & 44.23 & 41.34 \\
10K & 31.99 & 29.20 \\
15K & 26.43 & 25.36 \\
20K & 24.20 & 22.73 \\
25K & 22.05 & 20.83 \\
30K & 21.19 & 19.96 \\
\bottomrule
\end{tabular}
\end{tabular}
\caption{Training trajectory for 7B from-scratch models (seed 137). Left: OWT validation PPL. Right: WikiText-103 validation PPL. Thin keys lead throughout.}
\label{fig:7b_trajectory}
\end{figure}

\subsection{Experiment 7b: Extended Training at 7B Scale (20B Tokens)}\label{sec:exp7b}

\paragraph{Setup.}
We extend training to 20B tokens (3 epochs over OpenWebText, 305{,}175 steps), increasing the tokens-to-parameters ratio to $\sim$3.0. Architecture and hyperparameters are identical to Experiment~7.

\begin{table}[h]
\centering
\caption{7B LLaMA on OpenWebText (20B tokens), mean $\pm$ std over 2 seeds. Both architectures converge to the same quality; thin keys train 8\% faster.}
\label{tab:7b_extended}
\begin{tabular}{lrr}
\toprule
 & Full Attention & Thin Keys ($\dselect = \dmodel/4$) \\
\midrule
Parameters & 6.74B & 5.93B ($-$12\%) \\
OWT val PPL & $9.25 \pm 0.00$ & $\mathbf{9.24 \pm 0.00}$ ($-$0.1\%) \\
WT103 val PPL & $13.26 \pm 0.12$ & $\mathbf{13.00 \pm 0.07}$ ($-$2.0\%) \\
Wall-clock time & $261.5 \pm 0.1$h & $240.7 \pm 0.9$h ($-$8.0\%) \\
\bottomrule
\end{tabular}
\end{table}

\begin{figure}[t]
\centering
\includegraphics[width=\textwidth]{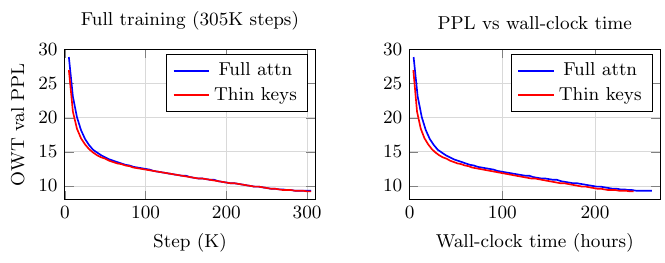}
\caption{Training trajectory for 7B models over 305K steps (20B tokens). \textbf{Left:} PPL vs step; thin keys lead early but converge by $\sim$150K steps. \textbf{Right:} PPL vs wall-clock time; thin keys reach any target PPL $\sim$20 hours sooner.}
\label{fig:7b_extended_trajectory}
\end{figure}

\paragraph{Results.}
At 20B tokens, both models converge to nearly identical perplexity ($9.24 \pm 0.00$ vs $9.25 \pm 0.00$ OWT, mean over 2 seeds), confirming the earlier advantage was a regularization effect. Thin keys are slightly better on WT-103 ($13.00$ vs $13.26$). Crucially, thin keys \emph{never fall behind}. Figure~\ref{fig:7b_extended_trajectory} shows PPL vs training step (curves merge by $\sim$150K steps) and vs wall-clock time (thin keys reach any target PPL $\sim$20 hours sooner). Downstream evaluation (Table~\ref{tab:c2_downstream}) confirms task parity: Hellaswag, ARC-Challenge, and WinoGrande scores match within noise ($<$1.5\%). Combined with 8\% speedup and 75\% key cache savings, thin keys are a strictly dominant choice at this scale.

\begin{table}[h]
\centering
\caption{Downstream evaluation of 7B from-scratch models (20B tokens, seed 137). Scores are low in absolute terms (OWT-only training) but thin keys matches full attention on all tasks.}
\label{tab:c2_downstream}
\begin{tabular}{lccc}
\toprule
Task & Full Attention & Thin Keys & $\Delta$ \\
\midrule
Hellaswag (acc\_norm) & 55.6 & 55.7 & $+$0.1 \\
ARC-Challenge (acc\_norm) & 29.8 & 29.4 & $-$0.4 \\
WinoGrande (acc) & 60.6 & 59.1 & $-$1.5 \\
MMLU (acc) & 23.4 & 23.3 & $-$0.1 \\
GSM8K (exact\_match) & 0.0 & 0.0 & 0.0 \\
\bottomrule
\end{tabular}
\end{table}

\paragraph{Comparison with GQA and MLA at 7B scale.}
Table~\ref{tab:kv_7b_analytical} compares thin keys against GQA and MLA analytically at the LLaMA-7B configuration ($\dmodel = 4096$, 128K context, bf16). GQA and MLA compress both K and V, achieving larger total savings (75--93\%) than thin keys alone (37.5\%). However, thin keys composes with GQA: applying $\dselect = \dmodel/4$ to GQA-8 yields 84.4\% total KV savings---approaching MLA's 93\% without learned up/down-projections or decoupled RoPE. MLA's joint latent means thin-keys composition is a no-op, but the thin-keys insight is \emph{already embedded} in MLA: DeepSeek-V2's $d_c = 512$ for 128 heads implies an effective per-head key dimension of 4, far below $\dhead = 128$. For practitioners, factored keys offer the simplest drop-in path for existing models; for new architectures, GQA + thin keys or MLA achieve more aggressive compression \citep{mha2mla2025}.

\begin{table}[h]
\centering
\caption{Analytical KV cache comparison at LLaMA-7B scale (128K context, bf16). MLA stores a shared latent $d_c$ plus a decoupled RoPE key ($d_h^R$). Thin keys compose with GQA by reducing $d_{\text{head}}^{QK}$ independently.}
\label{tab:kv_7b_analytical}
\begin{tabular}{lrrrr}
\toprule
Method & K cache (GB) & V cache (GB) & KV total (GB) & KV saved \\
\midrule
MHA (baseline) & 32.0 & 32.0 & 64.0 & --- \\
\midrule
Thin keys ($\dselect = \dmodel/4$) & 8.0 & 32.0 & 40.0 & 37.5\% \\
GQA-8 & 8.0 & 8.0 & 16.0 & 75.0\% \\
MLA ($d_c\!=\!512$, $d_h^R\!=\!64$) & \multicolumn{2}{c}{4.5 (joint)} & 4.5 & 93.0\% \\
\midrule
GQA-8 + thin keys & 2.0 & 8.0 & 10.0 & 84.4\% \\
\bottomrule
\end{tabular}
\end{table}

\subsection{Experiment 8: SVD + Fine-tuning at Scale (Mistral 7B)}

\paragraph{Setup.}
We apply SVD + QK fine-tuning to Mistral-7B (7.2B parameters, GQA with 32 query heads, 8 KV heads, $\dmodel = 4096$). We compress $W_K$ via SVD to ranks 512, 256, and 128, then fine-tune only QK projections on WikiText-103 for 3 epochs (10M tokens).

\begin{table}[h]
\centering
\caption{Mistral-7B: SVD compression + QK fine-tuning on WikiText-103.}
\label{tab:mistral7b}
\begin{tabular}{rrrrrr}
\toprule
Rank $r$ & Before FT & After FT & Control & vs Control & K cache saved \\
\midrule
1024 (none) & 6.69 & 5.91 & --- & baseline & 0\% \\
512 ($d_K/2$) & 6.81 (+1.7\%) & 5.93 & 5.91 & +0.3\% & 50\% \\
256 ($d_K/4$) & 7.84 (+17.1\%) & 6.03 & 5.91 & +2.0\% & 75\% \\
128 ($d_K/8$) & 12.34 (+84.4\%) & 6.10 & 5.91 & +3.2\% & 88\% \\
\bottomrule
\end{tabular}
\end{table}

\paragraph{Results.}
The pipeline scales to 7B with consistent quality (Table~\ref{tab:mistral7b}). At rank 256 (75\% K cache saved), the residual gap is +2.0\%---matching GPT-2 results exactly.

\paragraph{Downstream task evaluation.}
We evaluate on five benchmarks: Hellaswag (10-shot), ARC-Challenge (25-shot), WinoGrande (5-shot), MMLU (5-shot), and GSM8K (5-shot CoT). Table~\ref{tab:downstream} shows the results.

\begin{table}[h]
\centering
\caption{Downstream evaluation of SVD-compressed Mistral-7B. ``Ctrl+FT'' is the uncompressed model with identical fine-tuning.}
\label{tab:downstream}
\begin{tabular}{lccccccc}
\toprule
Task & Metric & Baseline & r512 +FT & r256 +FT & Ctrl+FT & $\Delta_{512}$ & $\Delta_{256}$ \\
\midrule
Hellaswag & acc\_norm & 81.2 & 81.4 & 80.7 & 81.3 & $+$0.1 & $-$0.7 \\
ARC-Challenge & acc\_norm & 54.0 & 54.1 & 53.4 & 54.4 & $-$0.5 & $-$1.7 \\
WinoGrande & acc & 75.4 & 73.2 & 72.1 & 73.2 & $+$0.0 & $-$1.6 \\
MMLU & acc & 60.1 & 55.2 & 54.4 & 55.7 & $-$0.9 & $-$2.3 \\
GSM8K & exact\_match & 38.4 & 27.7 & 25.8 & 29.9 & $-$7.4 & $-$13.7 \\
\bottomrule
\end{tabular}
\end{table}

At rank 512, compression is effectively lossless on knowledge and commonsense tasks ($<$1\% vs control). At rank 256, gaps remain modest (0.7--2.3\%) except for GSM8K, where multi-step math reasoning is more sensitive. Domain-matched fine-tuning recovers GSM8K quality: fine-tuning on GSM8K's own training split closes the compression gap from $-$13.7\% to just $-$1.2\% for r=256 (Section~\ref{app:gsm8k}). Generalization to held-out math benchmarks (Minerva Algebra, AGIEVAL AQuA-RAT) confirms transferable reasoning recovery with $<$2.5\% compression gaps.

\begin{table}[h]
\centering
\caption{Math generalization of GSM8K-fine-tuned models on held-out benchmarks.}
\label{tab:math_generalization}
\begin{tabular}{llccccc}
\toprule
Benchmark & Metric & Control & r512 & r256 & $\Delta_{512}$ & $\Delta_{256}$ \\
\midrule
GSM8K & exact\_match & 53.7 & 52.1 & 51.2 & $-$1.6 & $-$2.5 \\
Minerva Algebra & math\_verify & 14.2 & 14.1 & 12.4 & $-$0.2 & $-$1.9 \\
Minerva Pre-Algebra & math\_verify & 21.0 & 19.1 & 19.1 & $-$2.0 & $-$2.0 \\
AGIEVAL AQuA-RAT & acc & 15.7 & 17.3 & 17.3 & $+$1.6 & $+$1.6 \\
\bottomrule
\end{tabular}
\end{table}

% ============================================================
% 4. ANALYSIS
% ============================================================
\section{Analysis}

\subsection{Practical Benefit: KV Cache}

The primary practical benefit is inference-time KV cache reduction. For a representative 7B configuration ($\dmodel = 4096$, 32 layers, fp16), Table~\ref{tab:kvcache} shows savings at different context lengths.

\begin{table}[h]
\centering
\caption{KV cache memory per user ($\dmodel = 4096$, 32 layers, fp16).}
\label{tab:kvcache}
\begin{tabular}{lrrr}
\toprule
 & Standard & $\dselect = \dmodel/2$ & $\dselect = \dmodel/4$ \\
 & & (SVD, no retrain) & (train or SVD+FT) \\
\midrule
\multicolumn{4}{l}{\emph{128K context}} \\
\quad K cache & 33.6 GB & 16.8 GB & 8.4 GB \\
\quad V cache & 33.6 GB & 33.6 GB & 33.6 GB \\
\quad \textbf{Total} & \textbf{67.2 GB} & \textbf{50.4 GB} & \textbf{42.0 GB} \\
\quad Savings/user & --- & 16.8 GB (25\%) & 25.2 GB (37.5\%) \\
\midrule
\multicolumn{4}{l}{\emph{1M context}} \\
\quad Total & \textbf{524 GB} & \textbf{393 GB} & \textbf{328 GB} \\
\quad Savings/user & --- & 131 GB (25\%) & 196 GB (37.5\%) \\
\bottomrule
\end{tabular}
\end{table}

These savings compound with concurrent users and context length. Even the conservative SVD path saves 131\,GB per user at 1M context.

\subsection{Decode Throughput: Roofline Analysis}\label{sec:decode_throughput}

The KV cache savings translate directly to decode throughput gains.

\paragraph{Bandwidth model.}
Autoregressive decode is deeply memory-bandwidth-bound: on H100 SXM ($3.35$\,TB/s peak), each decode step reads model weights $W$ (shared across the batch) plus per-sequence KV cache.
For batch size $b$ and context length $n$, total bytes read per step are $W + b \cdot C_{\mathrm{kv}}$, where $C_{\mathrm{kv}} = 2\, L\, n_{\mathrm{kv}}\, d_{\mathrm{head}}\, n \cdot \beta$ and $\beta = 2$ for bf16.
Factored keys reduce \emph{both} $W$ (thinner $W_Q$, $W_K$ projections) and $C_{\mathrm{kv}}$ (thinner K cache).
The predicted speedup is:
\begin{equation}\label{eq:roofline}
\mathrm{Speedup}(b) \;=\; \frac{W + b \cdot C_{\mathrm{kv}}}{W' + b \cdot C'_{\mathrm{kv}}}
\end{equation}
This increases monotonically with $b$: at $b{=}0$ the gain comes only from smaller projections ($W/W'$); as $b \to \infty$ the cache term dominates and the speedup approaches $C_{\mathrm{kv}}/C'_{\mathrm{kv}}$.

\paragraph{Mistral-7B numbers.}
For Mistral-7B ($W{=}14.2$\,GB, $C_{\mathrm{kv}}{=}537$\,MB at $n{=}4096$), factored r256 gives $W'{=}13.2$\,GB (1.0\,GB saved from thinner Q/K projections) and $C'_{\mathrm{kv}}{=}336$\,MB (K cache shrunk $4\times$).
Table~\ref{tab:decode_throughput} reports both the bandwidth-model prediction from Eq.~\ref{eq:roofline} and measured throughput on a single H100 SXM.

\begin{table}[h]
\centering
\caption{Decode throughput for Mistral-7B with factored keys on H100 SXM (context 4096, 128 generated tokens). ``Predicted'' is the bandwidth roofline from Eq.~\ref{eq:roofline}.}
\label{tab:decode_throughput}
\small
\begin{tabular}{lcccccc}
\toprule
 & \multicolumn{5}{c}{Batch size} \\
\cmidrule(lr){2-6}
 & 1 & 4 & 8 & 16 & 32 \\
\midrule
\multicolumn{6}{l}{\emph{Measured throughput (tokens/s)}} \\
\quad Baseline ($d_k\!=\!128$) & 49.5 & 124.0 & 160.6 & 181.5 & 194.7 \\
\quad r512 ($d_k\!=\!64$)      & 53.5 & 142.9 & 192.8 & 223.2 & 243.0 \\
\quad r256 ($d_k\!=\!32$)      & 56.1 & 155.7 & 216.4 & 254.3 & 280.2 \\
\midrule
\multicolumn{6}{l}{\emph{Measured speedup}} \\
\quad r512 & 1.08$\times$ & 1.15$\times$ & 1.20$\times$ & 1.23$\times$ & 1.25$\times$ \\
\quad r256 & 1.13$\times$ & 1.26$\times$ & 1.35$\times$ & 1.40$\times$ & 1.44$\times$ \\
\midrule
\multicolumn{6}{l}{\emph{Predicted speedup (Eq.~\ref{eq:roofline})}} \\
\quad r512 & 1.06$\times$ & 1.08$\times$ & 1.10$\times$ & 1.14$\times$ & 1.19$\times$ \\
\quad r256 & 1.09$\times$ & 1.12$\times$ & 1.17$\times$ & 1.23$\times$ & 1.31$\times$ \\
\bottomrule
\end{tabular}
\end{table}

As predicted by the bandwidth model, measured speedups increase monotonically with batch size, since the KV cache fraction of total bandwidth grows from ${\sim}4\%$ at $b{=}1$ to ${\sim}55\%$ at $b{=}32$.
Measured speedups slightly exceed predictions (e.g.\ $1.44\times$ vs $1.31\times$ at $b{=}32$ for r256), likely due to improved cache locality from smaller tensors.
At batch size~1, the 14.2\,GB model-weight read dominates and KV savings are modest; at large batch sizes the cache term dominates and factored keys approach their theoretical maximum of $C_{\mathrm{kv}} / C'_{\mathrm{kv}} = 1.60\times$ for r256.
Standard SDPA suffices; no custom Flash Attention kernels are required.
Prefill roofline analysis is in Section~\ref{app:throughput}.

% ============================================================
% 5. RELATED WORK
% ============================================================
\section{Related Work}

\paragraph{Multi-Query, Grouped-Query, and Multi-Latent Attention.}
MQA \citep{shazeer2019fast} and GQA \citep{ainslie2023gqa} reduce KV cache by sharing KV heads across query groups. MLA \citep{liu2024deepseek} projects the joint KV state into a low-dimensional latent. These methods reduce head count or joint KV dimensionality; our approach reduces \emph{per-head key dimensionality} and composes with all three (Table~\ref{tab:kv_methods} in Section~\ref{app:arch_125m}). \citet{costoptimalgqa2025} show commonly used GQA configurations are suboptimal for long contexts.

\paragraph{Low-Rank Attention and KV Cache Compression.}
Linformer \citep{wang2020linformer} reduces the sequence dimension; we reduce the feature dimension. Loki \citep{loki2024} exploits low-rank key structure for sparse attention. LRQK \citep{lrqk2025} jointly decomposes QK matrices during prefill. SVD-LLM \citep{svdllm2025} applies truncation-aware SVD for post-training compression. KVQuant \citep{kvquant2024} and AsymKV \citep{asymkv2024} reduce bit width; ZACK \citep{zack2025} achieves zero-overhead dimensionality compression. \citet{compressionbarriers2025} prove $\Omega(\log n)$ space lower bounds for attention. Our dimensionality reduction is orthogonal to quantization and composes multiplicatively.

\paragraph{Efficient Attention.}
Flash Attention \citep{dao2022flashattention, flashattention32024} optimizes memory access patterns; sparse attention methods \citep{child2019generating, beltagy2020longformer, saap2025} reduce the attended set. These are complementary to reducing QK dimensionality.

% ============================================================
% 6. DISCUSSION
% ============================================================
\section{Discussion}

\paragraph{Limitations.}
We validate up to 7B parameters over 20B tokens (tokens-to-parameters $\sim$3). At truly Chinchilla-optimal budgets ($\sim$140B tokens), a modest cost may emerge as observed at smaller scales. While we evaluate perplexity and five downstream tasks, other capabilities (in-context learning, instruction following, long-context retrieval) may exhibit different sensitivity to QK dimensionality.

\paragraph{Flash Attention and composability.}
Most Flash Attention implementations assume $d_{\text{head}}^{QK} = d_{\text{head}}^{V}$, but standard SDPA already achieves 25--44\% decode speedups (Section~\ref{app:throughput})---decode is bandwidth-bound, so standard kernels suffice. Thin keys compose with KV quantization \citep{liu2024kivi, asymkv2024}: dimensionality reduction removes low-rank redundancy, then quantization compresses remaining elements, yielding up to $16\times$ combined key cache compression ($4\times$ from thin keys $\times$ $4\times$ from INT4).

\paragraph{Deployment paths.}
We identify three complementary paths, in order of increasing investment.
(1)~$K$-only SVD at $\dmodel/2$: 50\% key cache savings, zero retraining. (2)~SVD + QK fine-tuning to $\dmodel/4$: 75\% savings at $\sim$2\% cost. (3)~Training from scratch with thin keys for future architectures---analogous to how GQA was adopted into LLaMA-2 onward.

\section{Conclusion}

We identify a fundamental asymmetry in attention---selection requires only $\BigO(\log N)$ dimensions while value transfer needs full $\dmodel$---and derive \textbf{factored keys}, a zero-cost inference primitive that shrinks the key cache of any deployed model via SVD: no retraining, no calibration data, no prefill overhead. At 7B scale over 20B tokens (2 seeds), thin keys match full-attention quality ($9.24$ vs $9.25$ PPL) with task parity on downstream benchmarks, while training 8\% faster. For existing models, SVD + QK fine-tuning achieves 75\% key cache savings at $\sim$2\% cost across a 58$\times$ scale range (GPT-2 to Mistral-7B), with 25--44\% faster decode on H100. The approach composes with GQA and quantization for up to $16\times$ combined compression.

% ============================================================
% REFERENCES
% ============================================================

% ============================================================
% APPENDIX
% ============================================================
\section{Small-Scale Experiments}\label{app:small_scale}

Experiments 1--4 use a standard transformer decoder with pre-norm layer normalization, GELU activations, and learned positional embeddings. The only variable across configurations is $\dselect$.

\subsection{Experiment 1: Positional Selection (Copy-Back Task)}

\paragraph{Setup.}
We design a task where each token must copy the token from $K=8$ positions earlier: $y_t = x_{t-K}$. The source position is fixed regardless of token content, isolating purely \emph{positional} selection. Sequences of length 64 are generated with random tokens from a vocabulary of 16. The model uses $\dmodel = 64$, 4 heads, 2 layers.

\paragraph{Results.}
Table~\ref{tab:copyback} shows that even $\dselect = 4$ (1 dimension per head) achieves 100\% accuracy, though convergence is slightly slower. This confirms that positional selection---learning to attend to a fixed offset---requires minimal dimensionality.

\begin{table}[h]
\centering
\caption{Copy-back task: accuracy and convergence by $\dselect$.}
\label{tab:copyback}
\begin{tabular}{rrrr}
\toprule
$\dselect$ & $\dselect$/head & Best Accuracy & Converge Epoch \\
\midrule
4  & 1  & 100.0\% & 300 \\
8  & 2  & 100.0\% & 200 \\
16 & 4  & 100.0\% & 200 \\
32 & 8  & 100.0\% & 200 \\
64 & 16 & 100.0\% & 200 \\
\bottomrule
\end{tabular}
\end{table}

\subsection{Experiment 2: Content-Based Selection (Key-Value Retrieval)}

\paragraph{Setup.}
We construct sequences of 8 random key-value pairs from a vocabulary of 16 tokens, followed by a query key. The model must output the value associated with the queried key. Positions are randomized each batch, so positional information is useless; the model must match content. Architecture: $\dmodel = 64$, 4 heads, 4 layers.

\paragraph{Results.}
Table~\ref{tab:kvretrieval} reveals a sharp transition between $\dselect = 4$ and $\dselect = 8$. With 1~dimension per head, keys map to scalars and cannot be reliably separated via dot product, achieving only 65.2\% accuracy. With 2~dimensions per head, keys can point in distinct angular directions, enabling perfect matching. The $\BigO(\log_2 N)$ prediction gives a lower bound: 16~keys require $\log_2(16) = 4$ dimensions total, but the model needs a small constant factor above this minimum to learn reliable separation via gradient descent. In practice, $\dselect \approx 2\log_2 N$ appears sufficient.

\begin{table}[h]
\centering
\caption{Key-value retrieval: accuracy and convergence by $\dselect$.}
\label{tab:kvretrieval}
\begin{tabular}{rrrr}
\toprule
$\dselect$ & $\dselect$/head & Best Accuracy & Converge Epoch \\
\midrule
4  & 1  & 65.2\%  & did not converge \\
8  & 2  & 100.0\% & 1900 \\
16 & 4  & 100.0\% & 1900 \\
32 & 8  & 100.0\% & 1400 \\
64 & 16 & 100.0\% & 1000 \\
\bottomrule
\end{tabular}
\end{table}

\subsection{Experiment 3: WikiText-2 Language Modeling}

\paragraph{Setup.}
We train causal language models on WikiText-2 \citep{merity2017pointer} (~2M training tokens) with $\dmodel = 256$, 8 heads, 6 layers, $d_{\text{ff}} = 1024$, sweeping $\dselect \in \{8, 16, 32, 64, 128, 256\}$. Word-level tokenization with vocabulary size $\sim$29K ($\text{min\_freq} = 2$). Training uses AdamW with cosine learning rate schedule for 30 epochs.

\paragraph{Results.}
Table~\ref{tab:wt2} shows that $\dselect = 64$ ($\dmodel/4$) achieves a test perplexity of 122.24, essentially identical to the baseline of 122.22. Even $\dselect = 32$ loses only 1.3\%. Notably, $\dselect = 128$ \emph{outperforms} the baseline (120.76 vs 122.22), likely because reducing QK capacity acts as a regularizer in this heavily overfitting regime (train PPL 37 vs validation PPL 127).

\paragraph{Connecting to the $\BigO(\log N)$ prediction.}
The vocabulary contains $\sim$29K words, yet $\dselect = 64$ (8 dimensions per head) suffices. This implies that the effective number of selection patterns $N$ is far smaller than the vocabulary size. Intuitively, attention does not need to distinguish every individual word from every other---it needs to distinguish \emph{categories} of tokens relevant to different attention patterns: syntactic roles (subject vs.\ object vs.\ modifier), semantic clusters (topic-related words), positional patterns (recent vs.\ distant tokens), and punctuation or structural cues. The number of such categories is in the hundreds, not the tens of thousands. With 8 heads each operating in 8 dimensions, the model can represent $\sim\!2^8 = 256$ distinguishable patterns per head, which appears to be sufficient for the selection demands of language modeling. This is consistent with the observation that attention heads tend to specialize into a modest number of interpretable roles (positional, syntactic, rare-word) rather than implementing vocabulary-sized lookup tables.

\begin{table}[h]
\centering
\caption{WikiText-2 results. Model: $\dmodel = 256$, 8 heads, 6 layers. Baseline $\dselect = 256$.}
\label{tab:wt2}
\begin{tabular}{rrrrrrr}
\toprule
$\dselect$ & $\dselect$/head & Val PPL & Test PPL & $\Delta$PPL & QK Params & QK Saved \\
\midrule
8   & 1  & 133.78 & 126.48 & +3.5\%  & 24,672  & 97\% \\
16  & 2  & 132.67 & 125.49 & +2.7\%  & 49,344  & 94\% \\
32  & 4  & 130.51 & 123.78 & +1.3\%  & 98,688  & 87\% \\
64  & 8  & 129.34 & 122.24 & +0.0\%  & 197,376 & 75\% \\
128 & 16 & 126.42 & 120.76 & $-$1.2\% & 394,752 & 50\% \\
256 & 32 & 126.95 & 122.22 & --- & 789,504 & 0\% \\
\bottomrule
\end{tabular}
\end{table}

\subsection{Experiment 4: WikiText-103 Language Modeling}

\paragraph{Setup.}
To eliminate overfitting as a confound, we repeat the experiment on WikiText-103 \citep{merity2017pointer} (~100M training tokens) with the same architecture. Word-level tokenization with $\text{min\_freq} = 200$ yields a vocabulary of $\sim$22K. We train for 10 epochs with batch size 64 and 2000 warmup steps.

\paragraph{Results.}
Table~\ref{tab:wt103} shows the clean comparison. With 50$\times$ more training data, the model is capacity-limited (train PPL $\approx$ val PPL), eliminating the regularization confound. Here $\dselect = 128$ ($\dmodel/2$) incurs only a 2.1\% perplexity increase with 50\% QK savings, and $\dselect = 64$ ($\dmodel/4$) incurs 4.3\% for 75\% savings---larger than on WikiText-2, confirming that the WikiText-2 result was partly masked by overfitting. Nevertheless, the tradeoffs remain compelling: the relationship between $\dselect$ and perplexity is smooth and monotonic, allowing practitioners to choose their operating point on a clear Pareto frontier.

\begin{table}[h]
\centering
\caption{WikiText-103 results. Model: $\dmodel = 256$, 8 heads, 6 layers. Baseline $\dselect = 256$.}
\label{tab:wt103}
\begin{tabular}{rrrrrr}
\toprule
$\dselect$ & $\dselect$/head & Val PPL & Test PPL & $\Delta$PPL & QK Saved \\
\midrule
32  & 4  & 38.38 & ---   & +7.6\% & 87\% \\
64  & 8  & 37.22 & ---   & +4.3\% & 75\% \\
128 & 16 & 36.42 & 35.80 & +2.1\% & 50\% \\
256 & 32 & 35.67 & ---   & ---    & 0\% \\
\bottomrule
\end{tabular}
\end{table}

\section{Architecture Generalization at 125M Scale}\label{app:arch_125m}

\subsection{Experiment 6: Architecture Generalization (LLaMA 125M)}

\paragraph{Motivation.}
All training experiments so far use 10M-parameter vanilla transformers. We validate on a faithful LLaMA implementation at 125M parameters---12.5$\times$ larger.

\paragraph{Setup.}
We implement a standard LLaMA architecture with RMSNorm, SwiGLU FFN, Rotary Position Embeddings (RoPE), no bias terms, pre-norm residuals, and tied embeddings \citep{touvron2023llama}. The \emph{only} modification: $W_Q$ and $W_K$ project to $\dselect$ dimensions; $W_V$ remains $\dmodel$. When $\dselect = \dmodel$, the model is exactly standard LLaMA. Configuration: $\dmodel = 768$, 12 heads, 12 layers, $d_{\text{ff}} = 2048$, $\sim$101.7M parameters at baseline. Training: WikiText-103 with vocabulary truncated to 22K tokens (min frequency 200), 5 epochs, batch size 64, sequence length 512, cosine schedule with 2000-step warmup.

\begin{table}[h]
\centering
\caption{LLaMA 125M with asymmetric attention on WikiText-103. Comparison with 10M vanilla transformer from Experiment 4.}
\label{tab:llama}
\begin{tabular}{rrrrrr}
\toprule
$\dselect$ & $\dselect$/head & Params & Val PPL & $\Delta$PPL & QK saved \\
\midrule
768 (full) & 64 & 101.7M & 22.80 & --- & 0\% \\
192 ($\dmodel/4$) & 16 & 91.1M & 23.77 & +4.3\% & 75\% \\
96 ($\dmodel/8$) & 8 & 89.3M & 24.45 & +7.2\% & 87\% \\
48 ($\dmodel/16$) & 4 & 88.4M & 25.30 & +11.0\% & 94\% \\
\bottomrule
\end{tabular}
\end{table}

\paragraph{Results.}
The degradation ratios are remarkably consistent across architectures and scales:

\begin{center}
\begin{tabular}{lrr}
\toprule
$\dselect$ & 10M Vanilla (Exp.~4) & 125M LLaMA (Exp.~6) \\
\midrule
$\dmodel/4$ & +4.3\% & +4.3\% \\
$\dmodel/8$ & +7.6\% & +7.2\% \\
\bottomrule
\end{tabular}
\end{center}

The +4.3\% cost at $\dselect = \dmodel/4$ is identical across architectures despite a 12.5$\times$ scale difference and completely different design choices (LayerNorm vs RMSNorm, GELU vs SwiGLU, learned positions vs RoPE). This strongly suggests the QK dimensionality requirement is a fundamental property of the attention mechanism.

\paragraph{Comparison with alternative KV compression methods.}
We train the same 125M LLaMA architecture with Grouped-Query Attention (GQA) \citep{ainslie2023gqa} and Multi-Latent Attention (MLA) \citep{liu2024deepseek}. All models are trained from scratch with identical hyperparameters.

\begin{table}[h]
\centering
\caption{125M LLaMA: comparison of KV compression methods trained from scratch on WikiText-103. KV budget = per-token, per-layer cache size (K + V dimensions stored).}
\label{tab:kv_methods}
\begin{tabular}{llrrrr}
\toprule
Method & Config & Params & KV budget & KV saved & Test PPL \\
\midrule
MHA & 12 heads & 101.7M & 1536 & 0\% & 23.07 \\
\midrule
Thin keys & $\dselect = 384$ & 94.6M & 1152 & 25\% & 23.22 (+0.7\%) \\
Thin keys & $\dselect = 192$ & 91.1M & 960 & 37.5\% & 24.09 (+4.4\%) \\
\midrule
GQA & 6 KV heads & 94.6M & 768 & 50\% & 23.15 (+0.3\%) \\
GQA & 4 KV heads & 92.3M & 512 & 66.7\% & 23.32 (+1.1\%) \\
\midrule
MLA & $d_c = 768$ & 108.8M & 768 & 50\% & 23.11 (+0.2\%) \\
MLA & $d_c = 512$ & 101.7M & 512 & 66.7\% & 23.23 (+0.7\%) \\
\bottomrule
\end{tabular}
\end{table}

Table~\ref{tab:kv_methods} shows that all three approaches achieve strong compression with modest quality costs. Thin keys with $\dselect = 384$ achieves 23.22 test PPL---matching GQA-4 and MLA-512 quality despite compressing only keys. The practical implication is that thin keys \emph{compose} with GQA or MLA for further savings.

\section{Additional Analysis}\label{app:analysis}

\subsection{Scaling of Minimum $\dselect$}

Our experiments reveal a consistent pattern in the minimum $\dselect$ required for different selection tasks:

\begin{table}[h]
\centering
\caption{Minimum effective $\dselect$ scales with task complexity.}
\label{tab:scaling}
\begin{tabular}{lrrl}
\toprule
Task & $N_{\text{effective}}$ & Min $\dselect$/head & Prediction ($\log_2 N$) \\
\midrule
Positional (copy-back) & $\sim$10 offsets & 1 & $\log_2 10 \approx 3$ \\
Content (16 keys)      & 16 keys         & 2 & $\log_2 16 = 4$ (total) \\
Language (WikiText)     & $\sim$256 patterns & 8 & $\log_2 256 = 8$ \\
\bottomrule
\end{tabular}
\end{table}

The empirical results are consistent with the $\BigO(\log N)$ prediction. For language modeling, the effective $N$ appears to be approximately 256, suggesting that attention selection operates over a few hundred distinct semantic/syntactic categories rather than the full vocabulary space. This aligns with recent findings that key vectors naturally lie in a significantly lower-dimensional space than the model dimension \citep{loki2024}.

\subsection{Overfitting Masks the Effect}

The WikiText-2 vs WikiText-103 comparison reveals an important methodological point: on small datasets, reducing $\dselect$ can \emph{appear} costless (or even beneficial) because the model is overfitting. The WikiText-2 baseline has a train/val PPL ratio of 3.4$\times$, indicating massive overfitting. Removing QK capacity acts as implicit regularization. On WikiText-103, where the model is underfitting (train PPL $>$ val PPL), the true cost of reduced $\dselect$ becomes visible.

This suggests that results reported only on small benchmarks may overstate the losslessness of QK compression, and that large-scale experiments are essential.

\section{GSM8K Fine-tuning Progression}\label{app:gsm8k}

Table~\ref{tab:gsm8k_progression} shows the full progression of GSM8K recovery across fine-tuning experiments. Fine-tuning on out-of-domain data (WikiText-103, C4) yields GSM8K scores in the 22--32\% range with large compression gaps. Domain-matched fine-tuning on GSM8K's own training split (Experiment~F3, 7{,}473 chain-of-thought examples, $\sim$1.5M tokens) more than doubles scores and closes the compression gap for r=256 from $-$13.7\% to just $-$1.2\%.

\begin{table}[h]
\centering
\caption{GSM8K exact-match accuracy across fine-tuning experiments. All models use the same QK-only fine-tuning protocol (3 epochs, lr=$5 \times 10^{-5}$). ``Control'' = uncompressed model with identical fine-tuning.}
\label{tab:gsm8k_progression}
\begin{tabular}{llccccc}
\toprule
Exp & FT Data & Control & r512 & r256 & $\Delta_{512}$ & $\Delta_{256}$ \\
\midrule
-- & None (baseline) & 38.4 & 33.8 & 16.5 & -- & -- \\
A & WikiText-103 & 29.9 & 27.7 & 25.8 & $-$7.4 & $-$13.7 \\
F & C4 (10M tok) & 30.6 & 29.4 & 22.8 & $-$3.9 & $-$25.5 \\
F2 & C4 + Math (10M tok) & 31.7 & 28.2 & 24.1 & $-$3.5 & $-$7.6 \\
\midrule
F3 & GSM8K CoT (1.5M tok) & \textbf{53.7} & \textbf{52.5} & \textbf{52.0} & $\mathbf{-0.7}$ & $\mathbf{-1.2}$ \\
\bottomrule
\end{tabular}
\end{table}

This demonstrates that the GSM8K degradation was a fine-tuning data problem, not a fundamental compression limitation. Data quality and domain match matter far more than data volume: 1.5M tokens of in-domain CoT outperform 10M tokens of generic web text.

\section{Prefill Roofline and Flash Attention}\label{app:throughput}

\paragraph{Prefill roofline.}
Prefill differs from decode: the full $QK^\top$ product is computed over the entire prompt. At context length $s = 4096$ with Mistral-7B, a single layer's attention FLOPs are $\sim$137\,GFLOPs while KV reads are $\sim$2\,MB---yielding arithmetic intensity of $\sim$68{,}000~FLOP/byte, well above the H100's ridge point. Prefill attention is thus compute-bound. Reducing $d_k$ from 128 to 32 cuts $QK^\top$ FLOPs by $4\times$ per head. Standard FlashAttention-2 assumes $d_k = d_v$ for shared tile sizes; FlashAttention-3 \citep{flashattention32024} introduces more flexible tiling that could accommodate asymmetric dimensions. Standard SDPA with \texttt{enable\_math=True} already achieves 6--12\% prefill speedups.

\end{document}